\documentclass[letterpaper, 10 pt, conference]{ieeeconf} 
\IEEEoverridecommandlockouts                              %
\usepackage[pdftex]{graphicx}
\usepackage{url}
\usepackage{hyperref}

\usepackage[usenames, dvipsnames]{color}
\usepackage{soul}

\usepackage{mathptmx} %
\usepackage{times} %
\usepackage{amsmath} %
\usepackage{amssymb}  %
\usepackage{bm}
\usepackage[per-mode=symbol]{siunitx}
\usepackage{pdflscape}
\usepackage{outlines}
\usepackage{booktabs}

\makeatletter
\newcommand{\eqnum}{\leavevmode\hfill\refstepcounter{equation}\textup{\tagform@{\theequation}}}
\makeatother

\usepackage[backend=biber,      %
    style=ieee,
    bibstyle=ieee,              %
    sortcites=true,   
    mincitenames=1,
    maxcitenames=16,
    giveninits=true,            %
    backref=false
]{biblatex}
\renewcommand*{\bibfont}{\normalfont\small}
\usepackage{fancyhdr}
\newcommand{\mytitle}{\textbf{Submitted to appear in IEEE ICRA 2023.}}

\title{\LARGE \bf {
Towards Robust Autonomous Grasping with Reflexes Using High-Bandwidth Sensing and Actuation
}}

\author{Andrew SaLoutos$^{1}$, Hongmin Kim$^{1}$, Elijah Stanger-Jones$^{1}$, Menglong Guo$^{1}$, and Sangbae Kim$^{1}$
\thanks{$^{1}$Authors are with the Biomimetic Robotics Laboratory at the Department of Mechanical Engineering, Massachusetts Institute of Technology (MIT), Cambridge, MA, 02139, USA. {\tt\small\url{saloutos@mit.edu}}}
\thanks{This work was supported by the Advanced Robotics Lab of LG Electronics Co., Ltd. and the Toyota Research Institute (TRI).} }

\begin{document}
\bstctlcite{IEEEexample:BSTcontrol}

\maketitle
\thispagestyle{fancy}
\fancyhf{}		%
\fancyfoot[L]{\normalfont \sffamily  \scriptsize \mytitle}		%
\addtolength{\footskip}{-10pt}    %
\pagestyle{empty}

\begin{abstract}

Modern robotic manipulation systems fall short of human manipulation skills partly because they rely on closing feedback loops exclusively around vision data, which reduces system bandwidth and speed. 
By developing autonomous grasping reflexes that rely on high-bandwidth force, contact, and proximity data, the overall system speed and robustness can be increased while reducing reliance on vision data. 
We are developing a new system built around a low-inertia, high-speed arm with nimble fingers that combines a high-level trajectory planner operating at less than 1~\si{Hz} with low-level autonomous reflex controllers running upwards of 300~\si{Hz}. 
We characterize the reflex system by comparing the volume of the set of successful grasps for a naive baseline controller and variations of our reflexive grasping controller, finding that our controller expands the set of successful grasps by 55\% relative to the baseline.
We also deploy our reflexive grasping controller with a simple vision-based planner in an autonomous clutter clearing task, achieving a grasp success rate above 90\% while clearing over 100 items.

\end{abstract}

\section{Introduction} \label{sec:intro}

Achieving human-like versatility in robotic manipulation will depend on developing hands that are as nimble and reactive as human hands.
Much work has been done on developing taxonomies and design requirements for hands \cite{bullock2012hand, feix2015grasp}.
Still, state-of-the-art manipulation systems have not yet been able to replicate the human hand's functionality.
\par
Instead, many modern approaches rely on hardware initially intended for slow and precise tasks and deploy learning algorithms that depend on large amounts of vision data to carefully plan grasps \cite{fazeli2019see, gao2021kpam, zeng2018robotic}. 
These algorithms can plan the entire manipulation process, from arm motion down to fingertip contacts, but the high latency introduced by the vision systems results in grasping controllers that are unable to react while interacting with objects, which requires high control bandwidth. 
Even if the planning algorithms use contact and force data, the bandwidth of the vision system limits the execution speed and usually requires that the manipulation plan is quasi-static.
\par
We summarize the challenges faced by robotic manipulation systems as the ``last centimeter problem'', inspired by the ``last mile problem'' for delivery of goods.
In our ``last centimeter problem'', the unpredictability of contact with neighboring objects and the environment during the final stages of a grasp attempt and the risk of excessively disturbing the object to be grasped can impose harsh constraints on the entire manipulation plan, resulting in slow or conservative actions.
The dependence of many robotic manipulation systems on vision data exacerbates this challenge, as the cameras that record manipulation scenes often become occluded during the final stages of manipulation plans.
\par
\begin{figure}[t!]
    \centering
    \includegraphics[width=0.9\linewidth]{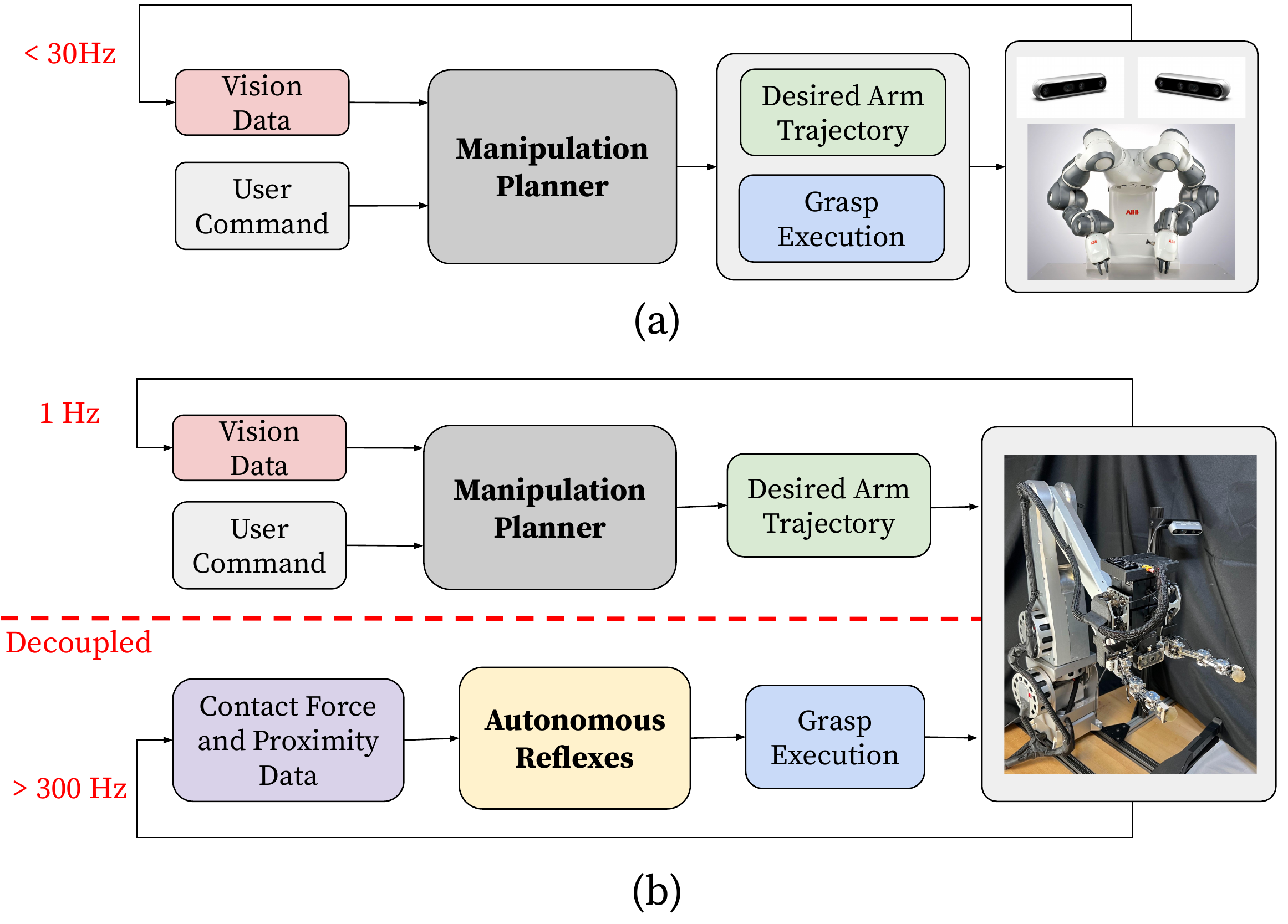}
    \caption{ \textbf{Manipulation control architectures.} Traditional, monolithic control architectures (a) are bottle-necked by the sampling frequency of the slowest sensor. Despite efforts to speed up frame rates and signal processing, bandwidths of vision-based control loops are typically limited to below 30~\si{Hz}. Our proposed control architecture (b) is decoupled into a slow, high-level planner and a layer of fast, autonomous reflexes that do not depend on vision data to achieve control bandwidths above 300~\si{Hz}.}
    \label{fig:arch_chart}
\vspace{-4mm}
\end{figure}

To address this problem, we propose a holistic method of designing manipulation systems while considering both hardware and controller requirements, starting from low-level, high-bandwidth behaviors, which we call reflexes.
For fast and robust manipulation, precise trajectory planning is insufficient: the real world is too messy and noisy and inevitably requires repetitive planning. 
Furthermore, given enough high-bandwidth robustness, precise planning is also not necessary.
To achieve this robustness, we focus on increasing the control bandwidth of a low-level, reflexive controller that is fully decoupled from a simple and imprecise high-level planner.
As reflexes are constructed to reason about contact interactions, finger motions, and potentially arm motions, the scope of a higher-level planner is reduced to reasoning only about the manipulation task. 
The reflexes layer in robustness and resilience through redundancy and are reminiscent of the subsumption architecture pioneered by~\textcite{brooks1986robust}. 
In our current system, we use these reflexes to close the grasping feedback loop locally in the hand without needing vision data or adding unnecessary planning complexity.
Fig.~\ref{fig:arch_chart} compares traditional manipulation system architectures with our proposed system.
\par

\begin{figure}
\centering
\includegraphics[width=0.8\linewidth]{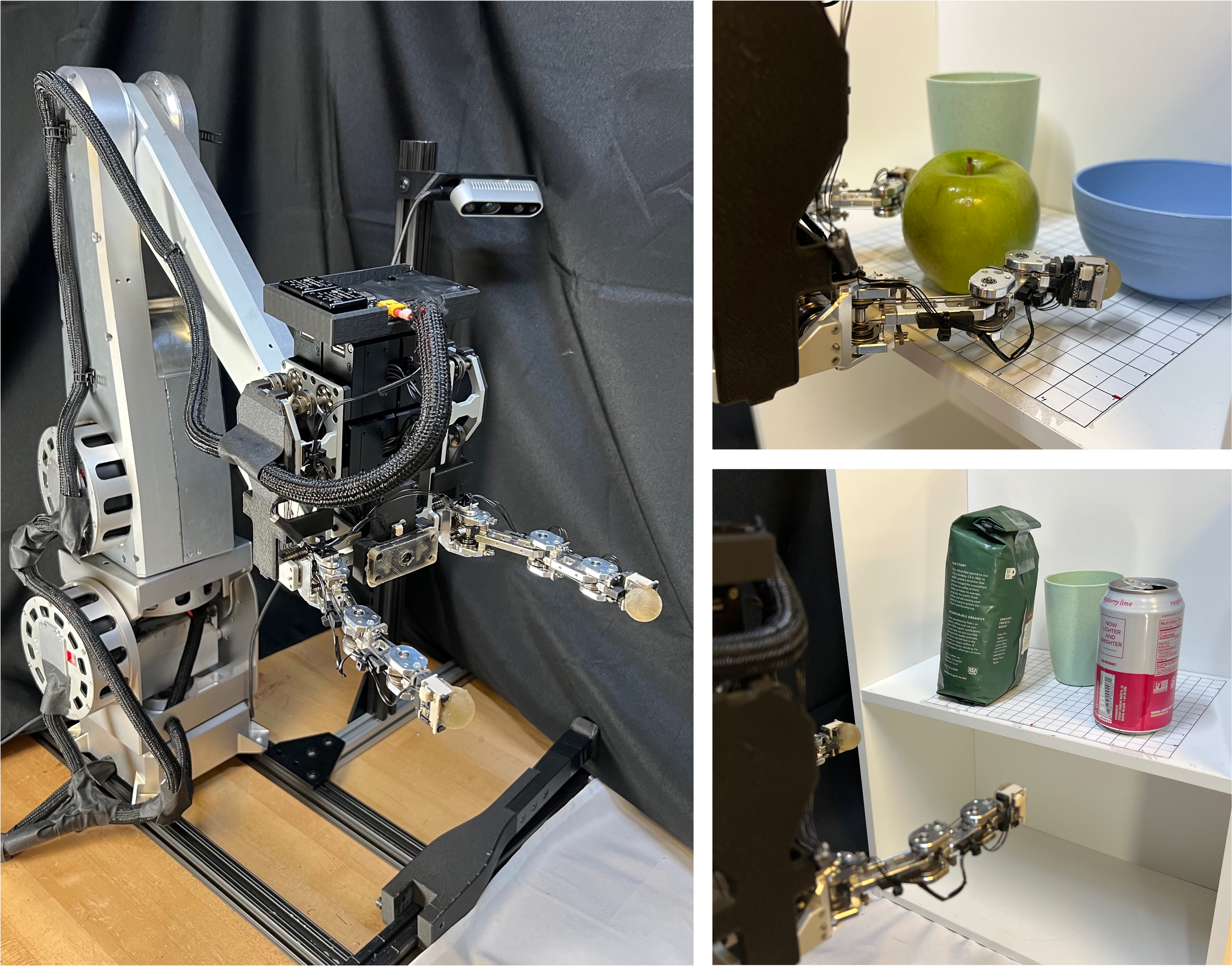}
\caption{ \textbf{Manipulation platform with reflexes for autonomous grasping.} Our reflexive grasping controllers utilize our manipulation platform's high-bandwidth actuation and low-latency sensing modalities to perform autonomous grasping tasks, such as clearing a cluttered shelf.
}
\label{fig:platform}
\vspace{-6mm}
\end{figure}

We introduce a reflexive manipulation algorithm for autonomous grasping and deploy it on our manipulation platform, which has high actuation bandwidth, dexterous fingers, and low-latency multimodal tactile sensors \cite{saloutos2022fast}.
To demonstrate the performance of the reflexes, we present two experiments. 
Our first experiment compares our reflexive grasping controller to a naive baseline grasp controller on a pick-and-place task. 
In our second experiment, we integrate our controller with a simple vision-based planner to complete an autonomous clutter-clearing task.

\section{Related Work} \label{sec:related}

Manipulation performance depends on the robot hardware and the control and planning systems. 
Our hardware is uniquely high-bandwidth, enabling faster performance than typical systems.
Our reflex-based grasping controller replaces low-level primitives, which can be simple routines or more complex reactive controllers, and can integrate with other planning algorithms, like those used in end-to-end grasping pipelines.

\subsection{Hardware for Manipulation}

Hardware platforms for grasping and manipulation are typically designed to be highly precise and are best suited for the factory floor.
Typical off-the-shelf robot arms, such as the UR5~\cite{ur5} or Kuka iiwa~\cite{kuka}, have high gear ratios and are not backdrivable. 
Instead, torque sensors are needed at each joint to close the force control loop, which limits the bandwidth and makes the systems stiff to collisions.
To achieve reflexive manipulation at human-level performance, we developed a fast proprioceptive manipulation platform~\cite{saloutos2022fast} based on the design principles developed for the Mini Cheetah~\cite{katz_mini}. 
The LIMS platform~\cite{kim2017anthropomorphic} achieves similar design goals; however, it uses a cable-driven design which significantly increases complexity. %
\par
Parallel-jaw grippers are commonly used in manipulation research~\cite{robotiq, ten_pas_2017, dong_2021}. 
These are simple to integrate; however, they are often not backdrivable and have limited or low-bandwidth force control. 
Robotic hands with more degrees of freedom, including the Shadow hand~\cite{shadowrobot}, Allegro hand~\cite{allegro}, and the DLR hand~\cite{butterfass2001dlr}, offer a wider range of capabilities but suffer from similar issues.
They are fragile and have low force control bandwidths, resulting in a poor ability to handle collisions.
Recently, \textcite{lin2021exploratory} and \textcite{bhatia2019direct} have studied reducing reflected inertia in grippers for impact-capable manipulation.
These design philosophies are critical for manipulation systems to operate in the real world, where they will face unexpected collisions and changes in the environment.
The hardware platform we present in this paper uses similar design principles applied to a higher degree-of-freedom system.
\par
Tactile sensing plays a vital role in the abilities of manipulation systems. 
Recently, the most common tactile sensors used have been vision-based systems such as Gelsight, Digit and the Soft-Bubble grippers~\cite{yuan2017gelsight, lepora2022digitac, alspach2019soft}.
These offer large amounts of data and are well suited to integration into learning pipelines.
However, they suffer from high latency, and a large amount of computation power is necessary to process the video stream. 
Traditional force-torque sensors offer low latency and high accuracy sensing but are too large for manipulation systems and suffer from acceleration-induced noise. 
We have presented a multimodal tactile sensor with low-latency force, contact location and proximity data~\cite{saloutos2022design}, which we use to enable our reflexive grasping controllers.

\subsection{End-to-end Software Pipelines for Grasping}

Manipulation systems are typically structured as end-to-end pipelines focusing on planning with vision data from both traditional cameras and vision-based tactile sensors. 
These pipelines are generally built with supervised learning or reinforcement learning (RL) algorithms~\cite{kalashnikov2018scalable}. 
While these systems are capable of a wide range of manipulation tasks~\cite{gao2021kpam}~\cite{dong_2021}, they face key bottlenecks when moving from the research lab to the real world.
Their planned trajectories often executed open-loop~\cite{ten_pas_2017}, which means they cannot adjust to environmental changes.
Even in systems that use closed-loop control, feedback is limited by camera frame rates and high processing times needed to parse the vision data, ultimately leading to end-to-end control bandwidths ranging from 5-50~\si{Hz}~\cite{morrison2018closing,ten_pas_learning_visuomotor,morrison2020learning}.
The bandwidth of the dynamics of the environments that these platforms interact with, mostly small and relatively lightweight objects, can be orders of magnitude higher than this, making it challenging to respond in real-time to unforeseen changes in the environment. 
This mismatch in dynamic bandwidth leads to systems operating quasi-statically to avoid disturbing the environment and taking as long as 20 to 60 seconds to complete a single grasp, which is significantly slower than a human. 
To avoid this mismatch, we are proposing a new structure built from fast and reactive reflexes that are not constrained by the bandwidth of a higher-level, vision-based planner.

\subsection{Reactive Grasping Controllers}

Reactive grasping controllers have been explored previously, using both tactile and proximity data to make quick adjustments to the system~\cite{teichmann2000reactive, reactive_gaifi}. 
The most common reaction during grasping is slip control, which typically uses a tactile sensor to determine when the object is slipping out of the hand~\cite{gunji2008grasping, cutkosky_slip}.
Additionally, reactions based on local analog proximity signals have been used to guide fingers around an object to ensure an enveloped grasp~\cite{shimojo_proximity}. 
In our prior work, we developed a slip detection reflex and an antipodal re-grasping reflex with a single degree-of-freedom gripper~\cite{saloutos2022fast} to achieve higher speed and robustness during teleoperated manipulation.
For these types of reflexes to work effectively, they require low-latency sensing and high-bandwidth force control at the gripper. 
In this work, we present several reflex designs in an integrated system that can handle various disturbances and scenarios.

\section{Manipulation Platform} \label{sec:hw}

Our manipulation system is shown in the left of Fig. \ref{fig:platform}. 
It consists of a low-inertia, high-speed arm and a dexterous two-finger gripper.
Each fingertip has a multimodal contact and proximity sensor.
\par
The arm has seven degrees of freedom: three at the shoulder, one at the elbow, and three at the wrist.
The three wrist joint axes intersect at the same point, creating a spherical joint.
To include the seventh degree of freedom at the wrist, we added a Dynamixel XM540 actuator to the wrist of the design presented in our previous work~\cite{saloutos2022fast}.
The additional degree of redundancy in the wrist pose is important for autonomous grasping as it allows for optimization of arm configurations given planned grasp locations or poses, as opposed to having at most one inverse kinematics solution for the desired grasp pose. 
\par
\begin{figure}[t]
\centering
\includegraphics[width=\linewidth]{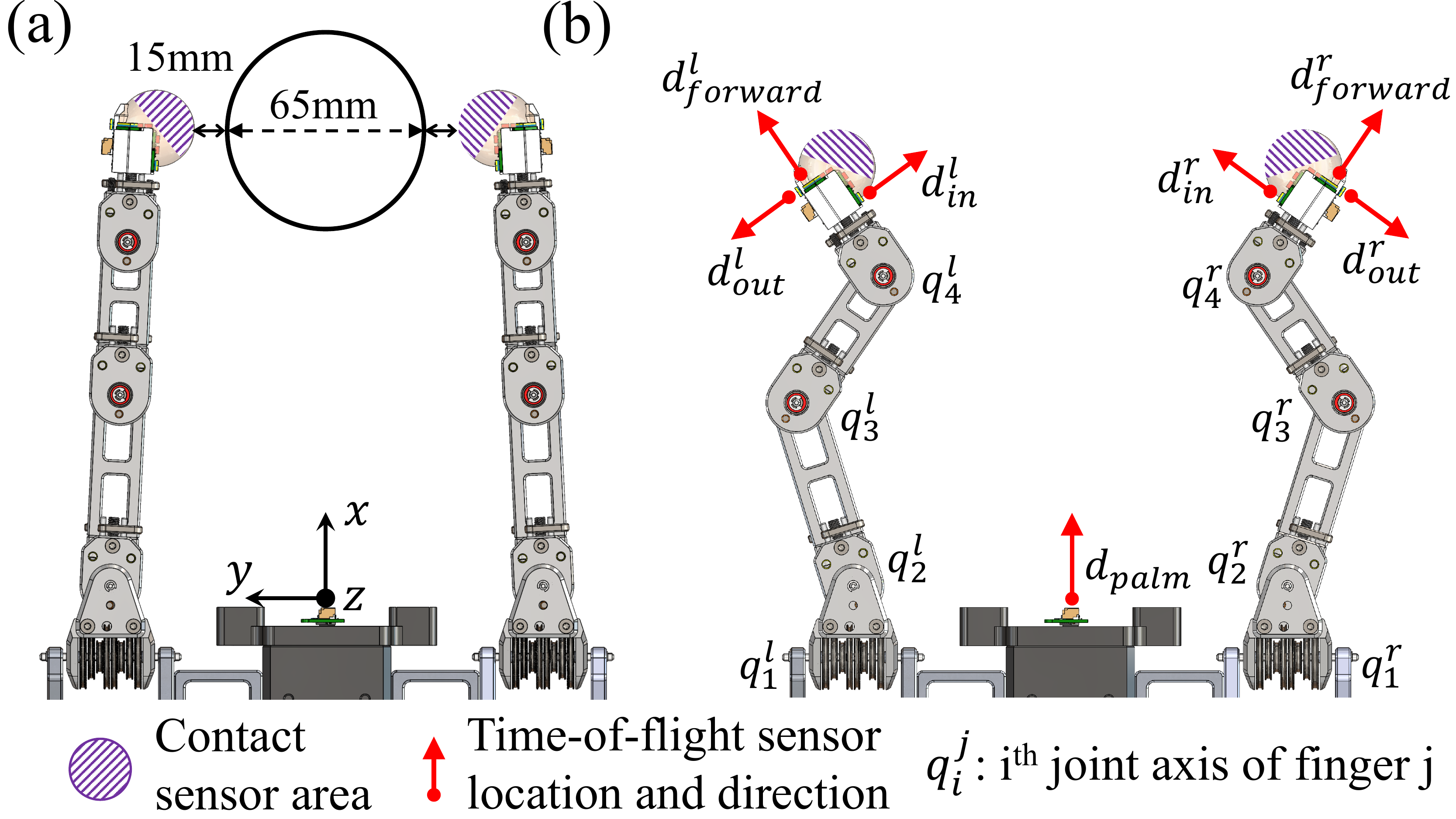}
\caption{ \textbf{Dexterous gripper with multimodal sensors.} Our system uses a two-fingered gripper with multimodal fingertip sensors. The contact sensor areas, time-of-flight proximity sensor directions, and joint axes are also labeled. (a) The nominal pose for the baseline grasping controller in Sec.~\ref{sec:exp1}. The finger width is set based on the cup diameter plus a clearance threshold. (b) The nominal pose for the reflex controllers in Sec.~\ref{sec:exp1}.}
\label{fig:hand_sensors}
\end{figure}

The gripper has two cable-driven fingers, each with four degrees of freedom.
Fig.~\ref{fig:hand_sensors} shows the fingers with labeled joint axes.
Each finger joint is driven by an antagonistic pair of tungsten cables routed through the finger to Dynamixel XM430 actuators.
All actuators are packed as close as possible to the axes of the wrist to minimize inertia while maintaining a wide range of motion.
While cable transmissions can become mechanically complex, using cables allows the fingers to be much slimmer and stronger than if actuators were placed directly at each joint.
The finger linkages are made of 7075-T6 aluminum alloy, with ball bearings supporting every joint shaft and routing pulley.
The fingers are lightweight and low-friction, enabling nimble manipulation actions.
The total gripper mass, including the wrist roll actuator, is approximately 1.2~\si{kg}.
We use a custom PCB for the gripper to receive commands from the control computer over a CAN bus, perform current control for the Dynamixel actuators, and send back gripper states and sensor information to the control computer.
\par
Each joint in the arm and fingers, represented by $q^i$, is torque-controlled using a proportional-derivative (PD) controller with a feedforward torque term:
\begin{equation}
    \tau^i_{command} = K^i_p (q^i_{des} - q^i) + K^i_d (\dot{q}^i_{des} - \dot{q}^i) + \tau^i_{ff}
\end{equation}
The control computer updates the desired positions, desired velocities, and torque commands at roughly 300~\si{Hz}.
The controllers for the arm actuators run at 1~\si{kHz}, and the controllers for the Dynamixels run at 500~\si{Hz}.
\par
Each fingertip has a multimodal sensor that can return contact force, location, and proximity data~\cite{saloutos2022design}, shown in Fig.~\ref{fig:hand_sensors}.
The sensor measures the 2-D contact location over its spherical surface, parameterized by the angles $\theta$ and $\phi$, and estimates the 3-D contact force at that location, $F = [f_x, f_y, f_z]$. 
The z-component of the contact force is the contact normal force, and the x- and y-components are the contact shear forces.
In addition to the contact data, the fingertip sensors capture pre-touch information from proximity data along three outward directions.
We also use a proximity sensor in the palm, and the vector of proximity data is given by:
\begin{equation} \label{eq:dist}
    d = [d^l_{out}, d^l_{forward}, d^l_{in},d_{palm},d^r_{in},d^r_{forward},d^r_{out}]
\end{equation}
The individual sensors are sampled at 200~\si{Hz}, with minimal time for processing overhead.

\section{Reflexive Grasping Algorithm} \label{sec:reflex}

\begin{figure}[t]
    \centering
    \includegraphics[width=0.9\linewidth]{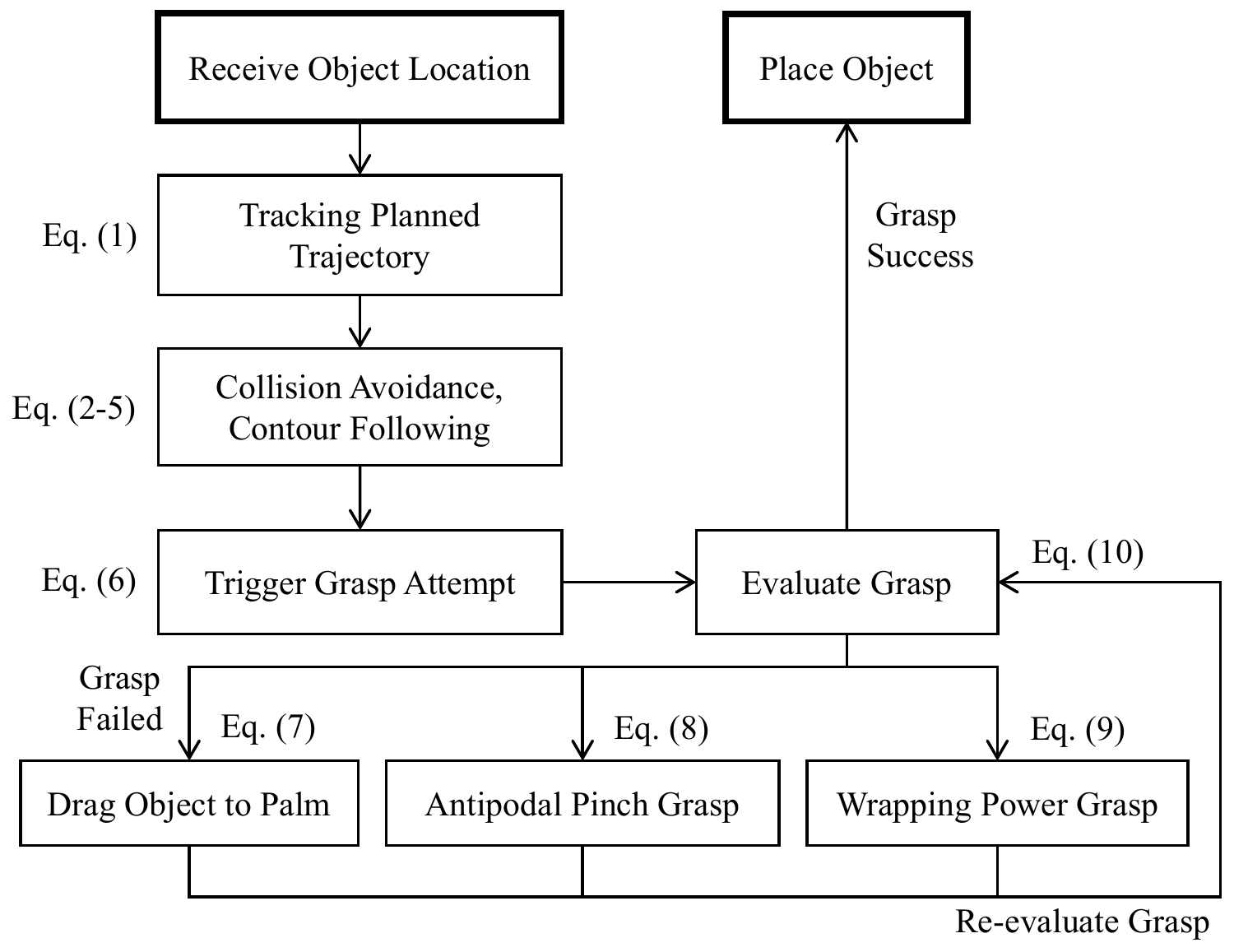}
    \caption{ \textbf{Reflexive grasping controller.} As the arm moves the gripper to the target location, the autonomous reflexes attempt to grasp the object. The high-level planner only needs to provide the target object location to the controller. }
    \label{fig:reflex}
\vspace{-6mm}
\end{figure}

Fig.~\ref{fig:reflex} shows a flowchart detailing our reflexive grasping controller.
The only high-level plan necessary for our controller is the current location of an object to be grasped and its desired final location. 
Once the controller receives a target grasping location, the arm moves the gripper towards the target, activating individual reflexes or combinations of reflexes along the way.
The reflexes use the arm and finger kinematics, force and proximity data, and contact kinematics to autonomously perform fast and robust grasping.

\subsection{Collision Avoidance and Contour Following} \label{sec:ref:coll_avoid}

As the hand approaches the target grasping location, the fingers hold a nominal ``open hand'' pose, shown in Fig.~\ref{fig:hand_sensors}.
The data from the proximity sensors are used to create virtual potential fields around the fingertips, with different stiffnesses and activation distances for each sensing direction from the fingertip: outwards, forwards, and inwards.
The sum of the forces from the individual potential fields yields a net force on the fingertip, given by:
\begin{equation}
\begin{aligned}
    \mathbf{F}^i &= \alpha_{out} K_{out} (d_{out} - d^{des}_{out}) \mathbf{e}_{out} \\
                 & \quad + \alpha_{forward} K_{forward} ( d_{forward} - d^{des}_{forward}) \mathbf{e}_{forward} \\
                 & \quad + \alpha_{in} K_{in} (d_{in} - d^{des}_{in}) \mathbf{e}_{in}
\end{aligned}
\end{equation}
where each $\alpha_i$ is an activation coefficient, each $\mathbf{e}_i$ is a unit vector along the corresponding direction from the fingertip, and stiffnesses and measured distances are represented by $K_i$ and $d_i$, respectively.
The coefficients $\alpha_i$ are determined by the threshold conditions:
\begin{equation}
\begin{aligned}
\alpha_{out} &= (d_{out} < d^{thresh}_{out}) \\
\alpha_{forward} &= (d_{forward} < d^{thresh}_{forward}) \\
\alpha_{in} &= (d_{in} < d^{thresh}_{in})
\end{aligned}
\end{equation}

\par
The virtual potential fields are activated as soon as the measured distance goes below the corresponding threshold.
For the outward and forward directions, the nominal distance for the potential fields, $d^{des}$, is the same as the activation threshold, $d^{thresh}$. 
For these directions, the forces on the fingertips are only repulsive, and enable reactive collision and obstacle avoidance. 
The potential fields due to the inward proximity sensors can be attractive and repulsive, as the activation threshold is not the nominal distance.
Once an object is within the activation threshold, the fingertip will attempt to maintain the desired distance from the object's surface, which yields a contour following behavior that allows the fingertips to react to object shapes and to passively wrap around objects while the hand is advancing to the target grasp position.
\begin{equation}
\begin{aligned}
    d^{des}_{out} &= d^{thresh}_{out} \\
    d^{des}_{forward} &= d^{thresh}_{forward} \\
    d^{des}_{in} &< d^{thresh}_{in} \\
\end{aligned}
\end{equation}

\begin{table}[t]
\caption{Algorithm Parameters}
\label{tab:alg_params}
\centering
\resizebox{\linewidth}{!}{
    \renewcommand{\arraystretch}{1.4}
    \begin{tabular}{ l  c  c } 
    \textbf{Purpose} & \textbf{Symbol} & \textbf{Value} \\
     \toprule
    Potential field distances & $( d^{thresh}_{out}, d^{thresh}_{forward}, d^{thresh}_{in}, d^{des}_{in} ) $ & $(9, 9, 9, 6)$$~\si{cm}$ \\ 
     \hline
    Potential field stiffnesses & $( K_{out}, K_{forward}, K_{in} ) $ & $(20, 30, 12)~\frac{\si{N}}{\si{m}}$ \\ 
     \hline
    Grasp triggering thresholds & $( d_{near}, d_{far}, d_{occlude} )$ & $(5, 9, 4)~\si{cm}$ \\ 
     \hline
    Antipodal angle threshold & $\gamma_{a}$ & $20\si{\degree}$  \\
     \hline
    Re-grasping radius threshold & $r_{power}$ & $3~\si{cm}$  \\
     \hline
    Grasp success thresholds & $( \gamma_v, \gamma_F )$ & $( 0.2\frac{\si{m}}{\si{s}}, 0.5~\si{N} )$  \\
     \hline
    Grasp termination time & $t_{fail}$ & $3.0~\si{s}$ \\ 
     \bottomrule
     \\
    \end{tabular}
}
\vspace{-6mm}
\end{table}

\subsection{Triggering and Evaluating Grasp Attempts} \label{sec:ref:trig_eval}

While the hand is moving towards the target grasp location, several conditions can trigger an early grasp attempt based on the proximity data and finger kinematics. 
These conditions are represented by $\beta_i$:
\begin{equation}
\begin{aligned}
    \beta_{near} &= (d_{palm} < d_{near}) \\
    \beta_{far} &= (d_{palm} < d_{far})  \\ %
    \beta_{tips} &= (q^l_{tip} < \theta_{close}) \land (q^r_{tip} < \theta_{close}) \\
    \beta_{occlude} &= (d^l_{forward} < d_{occlude}) \lor (d^r_{forward} < d_{occlude}) \\
\end{aligned}
\end{equation}
where $q^l_{tip}$ and $q^r_{tip}$ are the fingertip angles in the gripper frame.
The conditions $\beta_{near}$ and $\beta_{far}$ are based primarily on the palm proximity measurement and indicate that the object is between the fingers and close enough to the palm to be grasped.
With the condition $\beta_{tips}$, a grasp can be triggered if the fingertips have wrapped around the object.
Finally, for $\beta_{occlude}$, if the forward proximity measurement is too low, the finger is occluded and cannot proceed with the grasp. 
Since the controller can initiate a re-grasping attempt, it can rely on the re-grasping if a finger is blocked rather than causing the grasp attempt to fail. 
If the gripper reaches the final grasp target without a grasp already being triggered, a grasp attempt is initiated anyways.

\subsection{Re-grasping Decisions} \label{sec:ref:regrasp}

Once a grasp has been attempted, the controller determines if the grasp was successful or if a re-grasp needs to be attempted.
Using the measured contact frames, the controller approximates the object as a circle and estimates the radius and location in the gripper frame, $r_{obj}$ and $(x_{obj}, y_{obj}, z_{obj})$.
Based on the differences between the x-coordinates of the fingertip locations, $x^l_{tip}$ and $x^r_{tip}$, and the estimated object location, $x_{obj}$, in the gripper frame, the controller plans one of three re-grasps: \\

\begin{itemize}

\item If $x_{obj} > x^l_{tip}$ and $x_{obj} > x^r_{tip}$, the re-grasping trajectory pinches the object and pulls it closer to the palm. Then, the fingers are moved forward to achieve an antipodal pinch grasp, based on the new object location and the estimated radius, $r_{obj}$.\eqnum \\

\item If $x_{obj} < x^l_{tip}$ and $x_{obj} < x^r_{tip}$, the fingertips have successfully wrapped around the object, but the grasp has not been declared successful.
If the object radius is less than a threshold, $r_{obj} < r_{power}$, the controller plans an antipodal pinch grasp by moving the fingertips back towards the palm. \eqnum \\
            
\item Otherwise, the controller plans to extend the fingers to wrap around the object into a more secure power grasp.
If the signs of the differences between the fingertip x-coordinates and the object x-coordinate do not match, the controller also plans a wrapping power grasp. \eqnum \\
\end{itemize}
After the re-grasping trajectory is completed, the resulting grasp is evaluated again.

\subsection{Evaluating Grasp Attempts} \label{sec:ref:term}

Grasp attempts are evaluated using information from fingertip and palm proximity measurements, contact forces, and contact kinematics.
Once grasp has been triggered, the controller will continue to attempt the grasp until success or until a set amount of time, $t_{fail}$, has passed.
The controller declares a successful grasp when both fingertips have stopped moving, both contact normal forces are above a threshold, and the object is seen by the palm proximity sensor, represented as:
\begin{equation}
\begin{aligned}
    \beta_{success} &= (v^l_{tip} < \gamma_v) \land (v^r_{tip} < \gamma_v) \land \\
                     &(|F^l_z| > \gamma_F) \land (|F^r_z| > \gamma_F) \land \beta_{far}
\end{aligned}
\end{equation}

\section{Experiments} \label{sec:experiment}

\begin{figure}[t]
\centering
\includegraphics[width=0.95\linewidth]{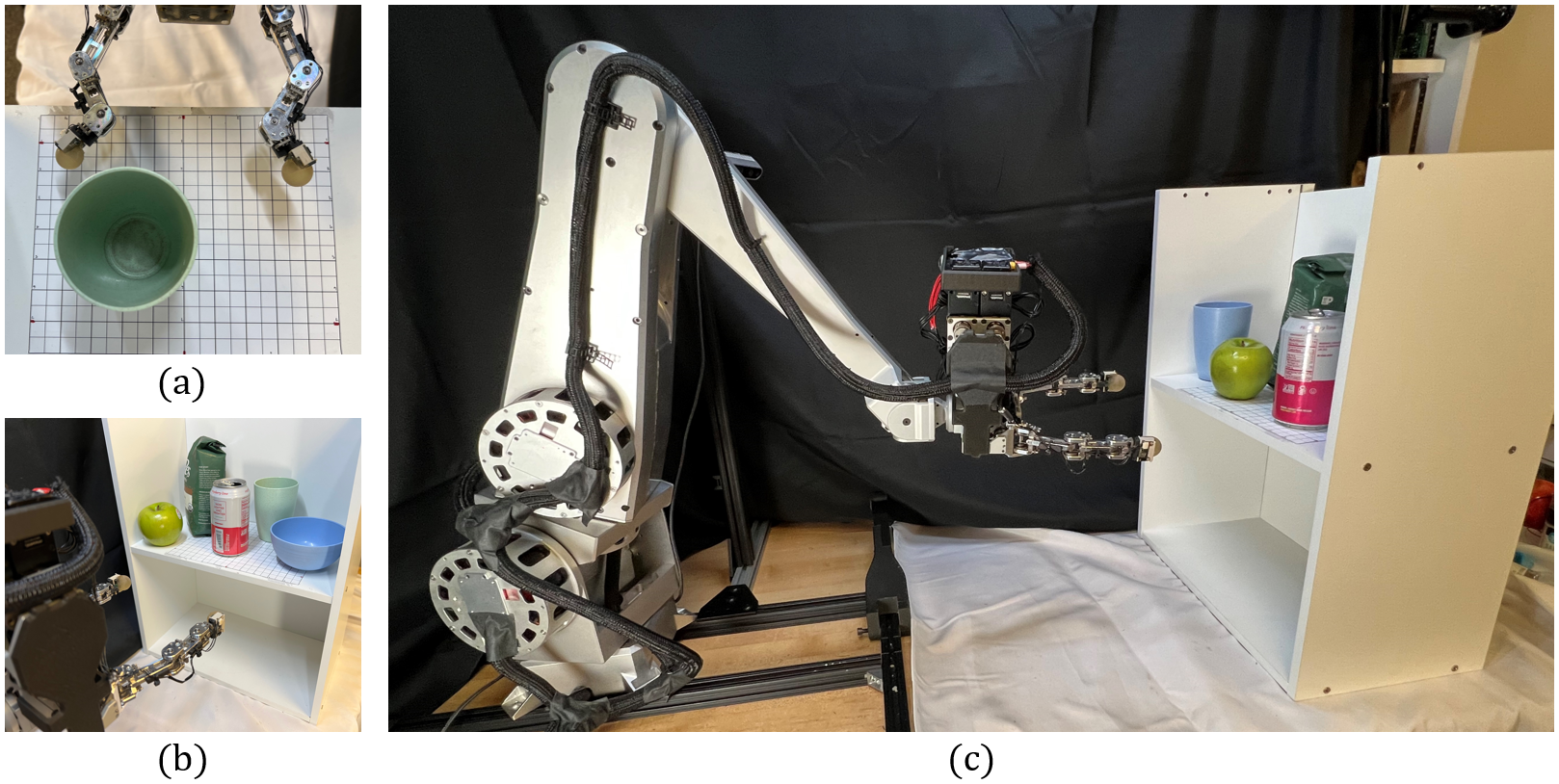}
\caption{\textbf{Experimental setups.} (a) The 12.5~\si{mm} grid and sample cup used to characterize the reflexive grasping controllers. (b) The variety of kitchen objects used for the autonomous clutter-clearing experiment. (c) The manipulation platform and shelf used for the experiiments.}
\label{fig:exp_setup}
\vspace{-4mm}
\end{figure}

\begin{figure*}[t] 
    \centering
    \includegraphics[width=\linewidth]{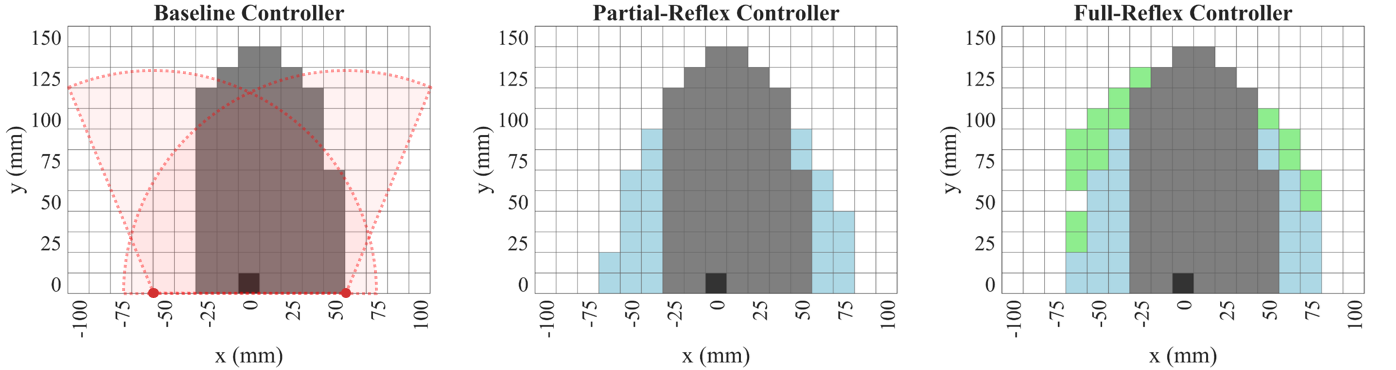}
    \caption{ \textbf{Characterizing the reflexive grasping controller.} As reflexes are added to the baseline grasp controller, the set of successful grasps expands. On the left-most plot, the base of each finger is marked by a red dot and the workspace is shown in red. The set of successful grasps for the baseline controller is shown in light grey on each grid plot, and the nominal grasp location is shown in dark grey. The expanded set of grid locations covered by the partial-reflex controller is shown in light blue, and the further expanded set of grid locations covered by the full-reflex controller is shown in green. The set of successful grasps for the baseline controller covers $11250~\si{mm}^2$. The total areas for the partial-reflex and full-reflex controllers are $14530~\si{mm}^2$ and $17500~\si{mm}^2$, respectively.}
    \label{fig:characterize}
\vspace{-4mm}
\end{figure*}

We present two grasping experiments using our reflexive grasping system.
In the first experiment, we characterize the difference in grasping capabilities between variations of our reflexive controller and a naive baseline controller.
In the second experiment, we use the reflexes to achieve robust autonomous grasping for a clutter-clearing task.

\subsection{Reflex Characterization} \label{sec:exp1}

For this experiment, we command the robot to repeatedly perform a pick-and-place task at a nominal grasp location.
To characterize the robustness of the grasping controllers, we record the set of successful grasps as the object is moved further and further from the nominal grasp location along a 12.5~\si{mm} grid. 
Displacing the object without changing the commanded grasp location isolates the effectiveness of the controllers from the measurement noise inherent to vision-based systems.
For a successful grasp, the gripper must grasp the object and then transport it to a predetermined ``place'' position. 
Fig.~\ref{fig:exp_setup}a shows the experiment setup with the fixed grid and the sample object, a cup that is roughly 110~\si{mm} tall and 65~\si{mm} in diameter.
\par
We characterize the performance of three grasping controllers.
The first controller is a naive baseline that performs a simple closing maneuver without any feedback information from the fingers or the fingertip sensors. 
The distance between the fingertips is chosen to be slightly wider than the cup diameter, as if the grasp command included the desired location and gripper width~\cite{morrison2018closing}.
\par
The second controller, called the partial-reflex controller, uses the collision avoidance and contour following reflexes but does not use the re-grasping reflex.
The third controller uses all of the reflexes, including re-grasping, so it is called the full-reflex controller.
For these controllers, the fingertips are spread wider apart to improve the effectiveness of the contour following.
The collision avoidance reflex mitigates any collisions that could happen during grasping with undesired objects or the environment due to the wider pose.
\par
Fig.~\ref{fig:characterize} shows the set of successful grasps for each of the three controllers.
Due to some inconsistencies in hardware initialization and joint frictions, the controller plots are not perfectly symmetric. Nevertheless, the overall trends are clear.
The contour following in the partial-reflex controller increases the width of the set near the nominal grasp location.
The re-grasping capabilities of the full-reflex controller are necessary to expand the set of successful grasps further away from the nominal location.
Based on the area covered by each controller, the set of successful grasps for the partial-reflex controller is 29.2\% larger than the set for the baseline controller.
The full-reflex controller achieves a 20.4\% increase relative to the partial-reflex controller and a 55.6\% increase relative to the baseline controller.
Each increase in the volume of the set of successful grasps can be interpreted as an increase in overall grasp robustness since the input to each controller is identical.

\subsection{Autonomous Grasping for Clutter-Clearing} \label{sec:exp2}

For this experiment, we combine our reflexive grasping controller with a vision algorithm for object identification to autonomously grasp objects from a cluttered shelf.
An Intel Realsense D435i camera is used to capture an image of the cluttered shelf, and the YOLOv7~\cite{wang2022yolov7} object classifier is used to detect graspable objects within the RGB image.
A point cloud is extracted from the depth image for the closest identified object.
Outlier points, defined by the top and bottom 10\% of depth measurements, are removed from the object point cloud, and the mean $(x,y,z)$ location of the remaining points in the world frame is sent as the target grasping location. 
During grasping, the desired wrist pose is set so that the gripper stays in the horizontal plane and that the inverse kinematics for the rest of the arm joint angles can be quickly solved analytically.
After startup, the only human involvement in the system is placing more objects into the environment.
The objects used, shown in Fig.~\ref{fig:exp_setup}b, include an apple, a bowl, a can, a bag of coffee grounds, and a cup.
\par
Table~\ref{tab:reflex} shows the grasping success rates for each object.
Across 117 total attempted grasps, the success rate is $90.6\%$.
While the bowl has by far the lowest grasp success rate at $68\%$, the other objects have an average grasp success rate of $94\%$. 
Across all of the successful grasps, the average time per grasp was 7.1 seconds, measured from the system receiving the estimated object location to the fingers releasing the object at the desired location.
This time included an average of 4.5 seconds for picking the objects, including the approach trajectory, and an average of 2.6 seconds for placing the objects. 
For context, the objects were placed approximately 35~\si{cm} away from the start of the grasping trajectory, which is within one human arm-length.

\begin{table}[t]
\caption{Grasping Success Rates}
\label{tab:reflex}
\centering
\begin{tabular}{l c c c} 
\bf{Object} & \bf{Trials} & \bf{Successes} & \bf{Success Rate}\\ 
 \toprule
 Apple & 26 & 25 & 96\%  \\ 
 \hline
 Coffee & 25 & 24 & 96\% \\ 
 \hline
 Cup & 26 & 24 & 92\% \\ 
 \hline
 Can & 24 & 22 & 92\% \\ 
 \hline
 Bowl & 16 & 11 & 68\% \\ 
 \hline
 \it{Total} & \it{117} & \it{106} & \it{90.6\%} \\ 
 \hline
\end{tabular}
\vspace{-4mm}
\end{table}

\section{Discussion and Conclusion} \label{sec:discuss}

Our experiments demonstrate improved robustness at three levels.
First, the reflexes improve grasping robustness to environmental uncertainties, such as noisy or occluded vision data, unstable objects, or unusual object shapes.
Second, the reflexes improve robustness to trajectory plans, allowing for simpler and lower-frequency planning.
This allows the planner to focus on high-level tasks instead of providing precise visual servoing.
Finally, our system achieves high-speed manipulation at a human scale with minimal impulses and collisions, which is critical for safe interaction that does not damage the environment or the robot.
\par
There are several improvements to be made to the system.
From an engineering perspective, the control frequency of the reflexes can be increased further with better system integration and low-level motor control for the Dynamixel actuators. 
We plan to push this loop frequency to at least 1~\si{kHz} to stabilize the fastest finger motions and further increase the reactive capabilities of the system. 
Furthermore, our current set of reflexes has been designed ad-hoc for everyday household objects, which fit very well into a relatively limited range of sizes and shapes.
In future work, we aim to expand the range of reflexes to account for broader classes of objects with different properties.
For example, softer objects, from a loaf of bread to a bag of chips, would require fundamentally different re-grasping approaches and grasp evaluation criteria, perhaps based on not damaging the objects or grabbing them by initially wiggling fingers underneath and then lifting.
In our current object set, the bowl has the lowest success rate by a wide margin. 
This is due to how the system approaches and tries to grasp it from the side, causing the fingers to push the bowl up and out of the grasp, rather than approaching from a vertical direction and pinching the side. 
There is a logical path forward for adding reflexes and capabilities, but currently the thresholds for contact forces, potential field distances, and grasping attempt triggers are all experimentally tuned.
We believe that building toward general manipulation capability, reliability, and flexibility will require layering many decoupled reflexes to address different scenarios in parallel instead of deploying a single do-it-all controller.
To effectively scale the system, learning approaches may be used to automate tuning and increase robustness across more object classes.
\par
We have shown a reflexive grasping algorithm that allows for robust autonomous grasping, even with a simple high-level planner.
The reflexes are designed to exploit our low-inertia manipulation platform and high-bandwidth sensors, and by decoupling the reflexive control from manipulation planning, the planner is able to operate at a much lower frequency than the reflex controllers: it is not responsible for solving the ``last centimeter'' problem.

\section*{ACKNOWLEDGMENTS}
The authors would like to thank Steve Heim for his thoughtful discussions and feedback.

\printbibliography

\end{document}